\documentclass[runningheads]{llncs}

\let\llncssubparagraph\subparagraph
\let\subparagraph\paragraph
\usepackage[compact]{titlesec}
\let\subparagraph\llncssubparagraph

\usepackage[hidelinks]{hyperref}
\usepackage{graphicx}
\usepackage{cite}
\usepackage{amsmath,amssymb,amsfonts}
\usepackage{algorithmic}
\usepackage{graphicx}
\usepackage{textcomp}
\usepackage{xcolor}
\usepackage{multirow}
\usepackage{booktabs}
\usepackage{array,multirow}
\usepackage{adjustbox}
\usepackage{titlesec}
\usepackage{amsmath}

\def\BibTeX{\rm B\kern-.05em{\sc i\kern-.025em b}\kern-.08em
    T\kern-.1667em\lower.7ex\hbox{E}\kern-.125emX}

\begin{document}

\title{Unlocking the Potential of Multiple BERT Models for Bangla Question Answering in NCTB Textbooks\
{\footnotesize \textsuperscript{}}
}

 \author{{ Abdullah Khondoker,   Enam Ahmed Taufik, Md Iftekhar Islam Tashik, S M Ishtiak mahmud, Antara Firoz Parsa\\}
 {Department of Computer Science and Engineering\\ School of Data and Sciences\\ 
Brac University\\}  
{\{abdullah.khondoker,  enam.ahmed.taufik, iftekhar.islam.tashik, sm.ishtiak.mahmud, antara.firuz.parsa\}@g.bracu.ac.bd}}

\authorrunning{}
\institute{}
\maketitle

\begin{abstract}
Evaluating text comprehension in educational settings is critical for understanding student performance and improving curricular effectiveness. This study investigates the capability of state-of-the-art language models—RoBERTa Base, Bangla-BERT, and BERT Base—in automatically assessing Bangla passage-based question-answering from the National Curriculum and Textbook Board (NCTB) textbooks for classes 6-10. A dataset of approximately 3,000 Bangla passage-based question-answering instances was compiled, and the models were evaluated using F1 Score and Exact Match (EM) metrics across various hyperparameter configurations. Our findings revealed that Bangla-BERT consistently outperformed the other models, achieving the highest F1 (0.75) and EM (0.53) scores, particularly with smaller batch sizes, the inclusion of stop words, and a moderate learning rate. In contrast, RoBERTa Base demonstrated the weakest performance, with the lowest F1 (0.19) and EM (0.27) scores under certain configurations. The results underscore the importance of fine-tuning hyperparameters for optimizing model performance and highlight the potential of machine learning models in evaluating text comprehension in educational contexts. However, limitations such as dataset size, spelling inconsistencies, and computational constraints emphasize the need for further research to enhance the robustness and applicability of these models. This study lays the groundwork for the future development of automated evaluation systems in educational institutions, providing critical insights into model performance in the context of Bangla text comprehension.

\end{abstract}

\section{Introduction}
The continuous evolution of Natural Language Processing (NLP) techniques has brought transformative advancements, particularly in the domain of Question Answering (QA) systems. QA systems are designed to interpret human language queries and provide relevant answers based on their understanding of the text. These systems have demonstrated significant progress in understanding and processing human language, enabling practical applications across various fields such as education, healthcare, and customer support. While some QA systems are open-domain, capable of addressing a wide variety of questions, others focus on more specific areas, known as closed-domain systems \cite{mervin2013overview}.

Despite these advancements, many languages, including Bengali, remain underrepresented in modern QA research, leaving a significant gap in resources and tools for these linguistic communities. Bengali, spoken by over 230 million people and ranked as the seventh most spoken language in the world \cite{das2022banglaser}, presents unique challenges for computational processing. Despite its widespread use, the language lacks the extensive NLP resources available to languages like English or Chinese. This scarcity of resources has led to a lag in the development of NLP tools for Bengali, particularly in specialized tasks such as question answering . Addressing these challenges is crucial to developing robust QA systems that can effectively cater to the needs of Bengali speakers and further integrate the language into modern AI applications.

This study embarks on a novel journey to harness the capabilities of state-of-the-art NLP models, focusing specifically on the Bengali language. Central to the objectives of this paper is the creation of a tailored dataset, meticulously curated to address specific goals. The dataset comprises approximately 3,000 passage-question-and-answer pairs drawn from the educational foundation of the Bengali language, covering classes six to ten. These pairs, carefully selected by human annotators in consultation with authoritative NCTB textbooks, provide a diverse and comprehensive QA resource. The dataset reflects various question types, ensuring its versatility for training and evaluation.

To process this expansive dataset, we adopted a systematic methodology leveraging the capabilities of three distinct models: BERT Base Multilingual Uncased \cite{bert_lr}, Bangla BERT \cite{banglabert_lr}, and RoBERTa \cite{roberta_lr}. Each model was subjected to rigorous tokenization, preprocessing, and training to optimize their performance in Bengali language comprehension. This comparative approach enables a deeper understanding of each model's strengths and limitations within the context of Bengali QA.

In recent years, advancements in NLP, particularly with transformer models like BERT \cite{bert_lr}, have revolutionized language understanding across many languages. BERT-Bangla \cite{banglabert_lr} has been a crucial tool for Bengali NLP, enabling improved performance in various tasks, including QA. However, when applied to closed-domain QA systems, challenges remain, particularly in handling domain-specific and complex cases \cite{bert_lr}. This motivates the current study, which aims to leverage the strengths of BERT-Bangla while addressing its limitations in the context of Bengali QA systems.

In the subsequent sections, we delve deeper into the dataset's nuances, elucidate the methodological approach, present our evaluation outcomes, and provide a comprehensive conclusion. The findings of this study contribute valuable insights into the potential of NLP models for underrepresented languages, paving the way for future advancements in automated evaluation systems. A sample diagram \ref{fig:fig1} for the system is given below:

\begin{figure}
\centering
\includegraphics[width=250px]{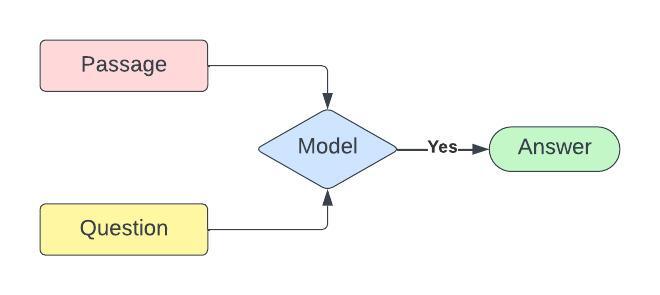}
\caption{Question Answering System} \label{fig:fig1}
\end{figure}

\section{Research Objectives}
This study aims to address the challenges in evaluating Bangla passage-based question-answering systems by focusing on the following key objectives:
\begin{itemize}
\item Evaluate the performance of state-of-the-art language models—RoBERTa Base, Bangla-BERT, and BERT Base—on Bangla passage-based question-answering tasks using metrics such as F1 Score and Exact Match (EM).
\item Investigate the impact of hyperparameter configurations, including batch size, learning rate, inclusion of stop words, and training epochs, on the performance of these models.
\item Identify and address the limitations of the dataset, such as spelling inconsistencies and the presence of non-factoid questions, which affect the accuracy and reliability of model evaluations.
\item Lay the foundation for developing robust, automated evaluation systems for Bangla text comprehension in educational institutions by analyzing model strengths and weaknesses.
\end{itemize}

\section{Literature review}

Banerjee et al. \cite{parsa_lr_1} paper pioneers the development of a factoid question-answering system tailored for Bengali, a low-resource language. This research represents an initial step in addressing the complex challenges inherent in constructing such a system. The authors propose a system capable of answering natural language questions in a human-like manner. Questions are classified into five categories, with the system primarily focusing on factoid questions in three distinct domains. Challenges in developing question-answering systems for low-resource languages, such as Bengali, include the prevalence and diverse positions of interrogatives and the scarcity of language processing tools. BFQA, their Factoid QA system for Bengali, comprises a pipeline architecture with three key components: question analysis, sentence extraction, and answer extraction. The authors introduce several question analysis processors, including QType and Expected Answer Type identification, named entity recognition, question topical target identification, and keyword identification. Evaluation employs the Mean Reciprocal Rank (MRR) metric, yielding an MRR of 0.32, albeit with varying performance across domains. Despite the achievement in advancing Bengali factoid question answering, this research acknowledges the system's lower accuracy compared to European languages, attributed to underperforming components like the shallow parser and named entity recognition system. In summary, this work contributes significantly to low-resource language question-answering while recognizing ongoing challenges.

In the paper by Zhou et al. \cite{parsa_lr_2}, critical issues of overconfidence and over-sensitivity in existing Reading Comprehension (RC) models are effectively addressed. Through their experiments, the study showcases significant improvements in the robustness of RC models. The key innovation of the paper lies in the development of a method that incorporates external knowledge to enforce a range of linguistic constraints, including entity, lexical, and predicate constraints. This integration empowers the model to generate more precise predictions, not only for semantic differences but also for semantic equivalences in adversarial examples. Furthermore, Zhou et al.\cite{parsa_lr_2} introduce posterior regularization into RC models, a crucial addition that contributes to enhancing the resilience of the underlying RC models. By seamlessly incorporating linguistic constraints and posterior regularization into the learning process, the proposed method not only bolsters the robustness of base RC models but also successfully integrates these constraints, resulting in more accurate and adaptable RC systems.

Bechet et al. \cite{tashik_lr_1} embark on an intriguing exploration of question-answering within digitized archive collections, with a particular focus on its applicability to Social Science studies. Their research introduces an innovative approach that harnesses the power of a BART Transformer-based generative model, enhanced by semantic constraints, to tackle the task of question generation. This pioneering approach holds great promise in the field of natural language processing, especially within the niche of archive collection studies. The study meticulously conducts experiments on three distinct corpora: FQUAD, CALOR-QUEST, and ARCHIVAL, each presenting its own set of complexities and challenges. The findings from these experiments unveil a notable enhancement in the quality of questions generated when semantic annotations are incorporated, underscoring the effectiveness of this approach in improving question generation. However, an intriguing observation emerges during the evaluation of Machine Reading for Question Answering (MRQA) performance. When models trained on automatically generated questions are used for evaluation, the study notes no significant improvements, particularly on the CALOR-QUEST and ARCHIVAL datasets. This intriguing finding underscores the nuances associated with using automatically generated questions in the evaluation process, opening up avenues for further investigation in this domain. Nevertheless, the research shines in its performance on the formidable ARCHIVAL dataset, demonstrating the robustness and adaptability of the proposed approach, even in challenging contexts. The analysis thoughtfully highlights the distinctions between questions generated by experts and those sourced from crowdsourcing efforts, shedding light on the importance of question quality in this context.\cite{tashik_lr_1}

Manav et al. \cite{tashik_lr_2}, to address the lack of a Bangla Question/Answer Dataset, introduces BanglaRQA, a reading comprehension-based question-answering dataset for Bangla. It contains 3,000 context passages and 14,889 question-answer pairs, covering answerable and unanswerable questions across four question categories and three answer types. The paper \cite{tashik_lr_2} also evaluates four Transformer models on BanglaRQA, with the best model achieving 62.42 percent EM, and 78.11 percent F1 scores. However, further analysis reveals variation in performance across question-answer types, indicating room for improvement. The paper \cite{tashik_lr_2} demonstrates the effectiveness of BanglaRQA as a training resource by achieving strong results on the bn\_squad dataset.

Devlin et al. \cite{nihal_lr_1} explore the application of NLP techniques, specifically neural language models, for generating question/answer exercises from English texts. The aim is to support ESL teaching to children by generating beginner-level exercises. The proposed approach involves a four-stage pipeline: pre-processing, answer candidate selection, question generation using the T5 transformer-based model, and post-processing. Evaluation of benchmark datasets demonstrates comparable results to previous works. However, limitations are identified, such as imperfect co-reference resolution and errors in question generation. Future work involves refining the system by exploring other language models, expanding the evaluation corpus, and fine-tuning the model for specific English proficiency levels. The developed tool has the potential for integration into an educational platform for English language teaching. \cite{nihal_lr_1}

Liu et al. \cite{nihal_lr_2} investigate the development of a deep learning-based question-answering system in Bengali, aiming to overcome the limitations and lack of progress in this field. By leveraging state-of-the-art transformer models, the research focuses on training a QA system using a synthetic reading comprehension dataset translated from SQuAD 2.0. Furthermore, a human-annotated QA dataset sourced from Bengali Wikipedia is utilized for evaluating the models. Comparative analysis with human children provides valuable insights and establishes a benchmark score. The research  \cite{nihal_lr_2}  emphasizes the importance of addressing the challenges in low-resource language settings, particularly in the context of reading comprehension-based question answering.

In this paper Bhattacharjee et al. \cite{abdullah_lr_1} introduce the task of Multi-Question Generation, aiming to generate diverse questions assessing the same concept. It addresses the limitation of existing systems that generate only one question per answer. The paper proposes an evaluation framework based on desirable question qualities and presents results comparing different question generation approaches.  The authors highlight the issue of word overlap between generated questions and input passages and propose metrics to measure question answerability, semantic similarity, and distinct wordings. Future work includes exploring human evaluation metrics, reinforcement learning objectives, and advanced paraphrase systems. The paper suggests incorporating teacher evaluation to define desirable question properties and evaluate the educational impact of diverse question wordings. The publicly released pipeline holds the potential for enriching educational resources at scale.\cite{abdullah_lr_1}

\section{Dataset}
In the context of our project's focus on the Bangla language, a pivotal component involves the creation of a dataset tailored to our objectives. Emphasizing the essence of customization, we undertook the task of dataset construction, marking a foundational stride in the development of an effective Bangla Question-Answering system. Our dataset \cite{dataset_lr} is a result of a meticulous curation process involving around 3,000 question-and-answer pairs. These pairs, meticulously selected by human annotators consulting NCTB textbooks from classes six to ten, offer a contextual and informative foundation. Each passage in the dataset, averaging 387 words, provides substantial context for the subsequent question answering. Importantly, human annotators meticulously collected responses for each question type, ensuring the dataset's reliability and relevance. The primary objective driving this endeavour is the development of a proficient Bangla question-answering system. To this end, we painstakingly organized the dataset into training and validation subsets, with each subset elegantly encapsulated within CSV files. These files harmoniously interweave multiple passages with their corresponding questions and expertly annotated answers. By taking the reins in dataset creation, we lay the groundwork for a question-answering system deeply rooted in precision, contextual understanding, and linguistic nuance intrinsic to the Bangla language. Here [~\ref{fig: Dataset}] is an example of our dataset: 

\begin{figure}[htbp] 
\centering
\includegraphics[width=\textwidth]{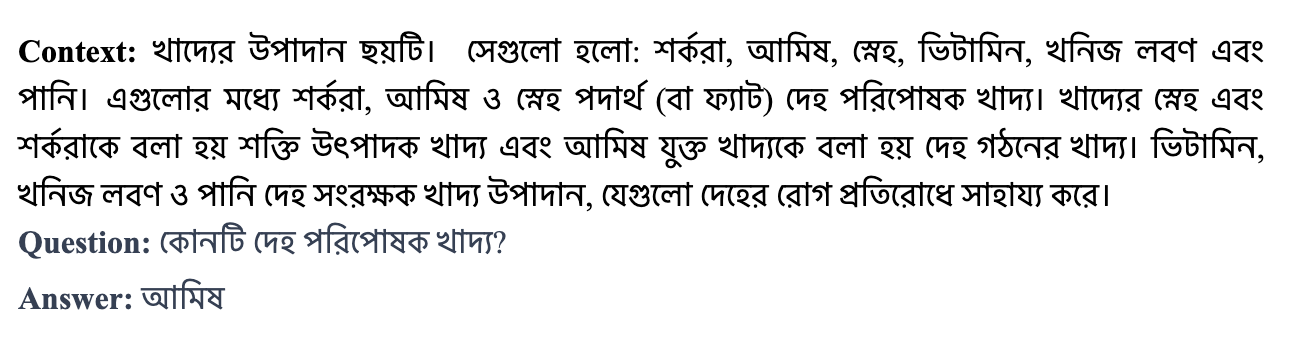}
\caption{Sample of Dataset} \label{fig: Dataset}
\end{figure}

\section{Methodology}
\begin{figure}
\centering
\includegraphics[width=\textwidth]{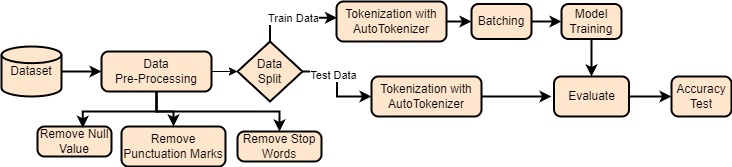}
\caption{Q\&A Model Process Flowchart} \label{fig:Workflow}
\end{figure}

This segment outlines the framework for an Automated Question Answering System in Bengali. Given the limited prior work in Bangla question answering, our objective is to develop a system that provides accurate responses to user queries, addressing the gap in this area. In this research endeavour, we meticulously processed the Bangla question-answering dataset to prepare it for subsequent modelling and evaluation. A step-by-step approach was undertaken to ensure the quality and integrity of the dataset. The model represented in [~\ref{fig:Workflow}] of our workflow-

\subsection{Data Preprocessing}
The initial stage of data preprocessing involved the removal of null values from the dataset, ensuring that the foundational dataset was devoid of any missing or incomplete entries. Following this step, we employed a Basic Tokenizer to facilitate the creation of tokenized data. The Average number of tokens is shown in [~\ref{fig:tokens}]:

\begin{figure}
\centering
\includegraphics[width=250px]{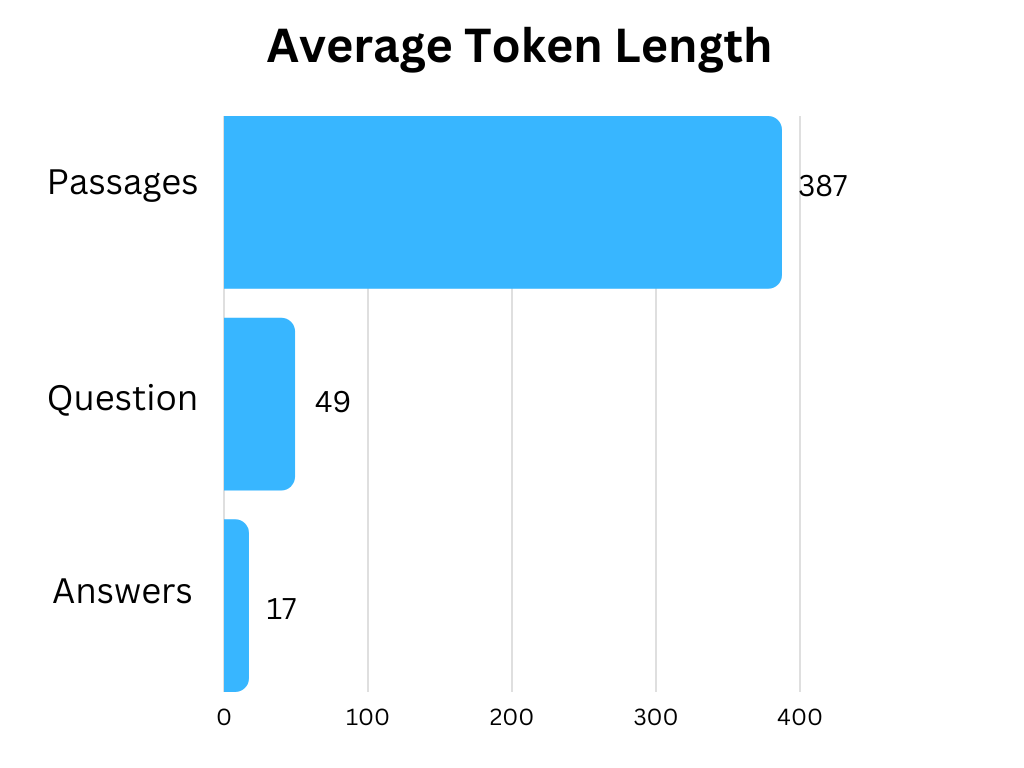}
\caption{Average token size of Passage, Question \& Answer} \label{fig:tokens}
\end{figure}

This tokenization procedure was instrumental in the removal of punctuation marks (e.g.  \$, \%, \#, *, -, etc.) and stop words\cite{stopwords_bn_lr}, streamlining the text data for subsequent analyses. Subsequently, we restored the dataset to its original format, preserving its coherence.
To accurately identify the span of answers within the provided passages, we meticulously calculated the start and end indices of the answers. The answer column was then transformed into a structured dictionary, encompassing the answer itself along with its corresponding start and end indices.

\subsection{Dataset Splitting}
The dataset was divided into distinct training and testing subsets to facilitate the model development and evaluation process. Specifically, a 70\%-30\% train-test split was employed, ensuring a robust assessment of model generalization.

\subsection{Tokenization with AutoTokenizer}
Our empirical findings revealed that, among the considered tokenization techniques, AutoTokenizer exhibited superior performance during model training. This tokenizer, with its adaptive tokenization strategy, proved to be particularly adept at handling the intricacies of Bangla text. To equip the dataset for subsequent modelling endeavours, we harnessed the power of AutoTokenizer. The AutoTokenizer provided by the Hugging Face Transformers library was employed for tokenization. It tokenizes both the questions and contexts, breaking them down into subword units and mapping them to corresponding token IDs. Additionally, attention masks were generated to indicate which tokens should receive attention during training. After tokenization, the data was preprocessed to generate inputs that the model could process. These inputs included tokenized sequences, attention masks, and answer span positions. The resulting tokenized inputs were organized into a structured format, which included input IDs, attention masks, start positions, and end positions for answer spans. The tokenized inputs were then transformed into a list of dictionaries, where each dictionary contained input IDs, attention masks, start positions, and end positions. These dictionaries were structured to facilitate feeding the data into the model during training. This tokenization technique incorporated essential parameters, including truncation (enabled), padding (set to 'max-length'), maximum token length (512 tokens), and the return of attention masks and tokenized tensors (return-attention-mask=True, return-tensors='pt'). Notably, we refrained from adding any special tokens to preserve the native structure of the Bangla text.

\begin{figure}[htbp] 
\centering
\includegraphics[width=350px]{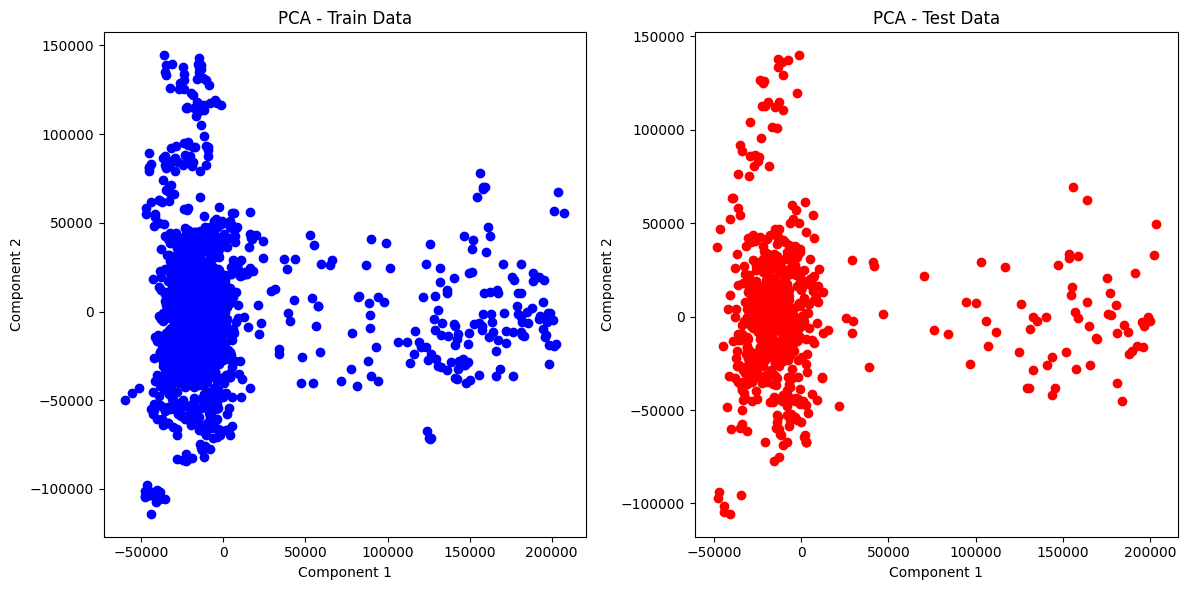}
\caption{PCA - Train \& Test for RoBERTa Base} \label{roberta}

\centering
\includegraphics[width=350px]{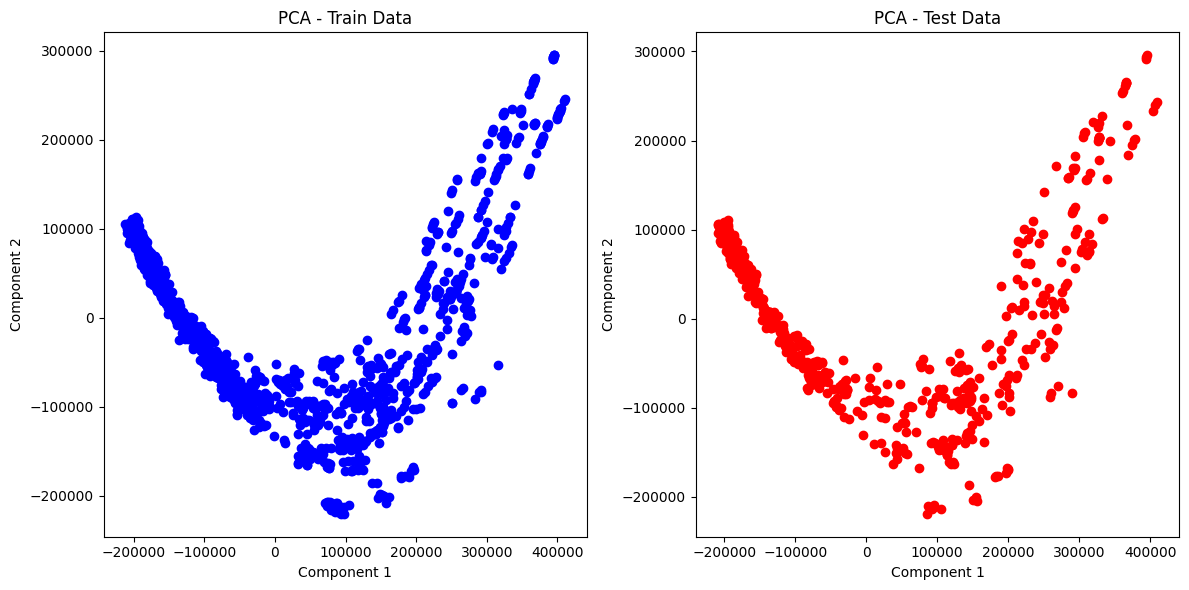}
\caption{PCA - Train \& Test for BERT Base} \label{bert_base}

\centering
\includegraphics[width=350px]{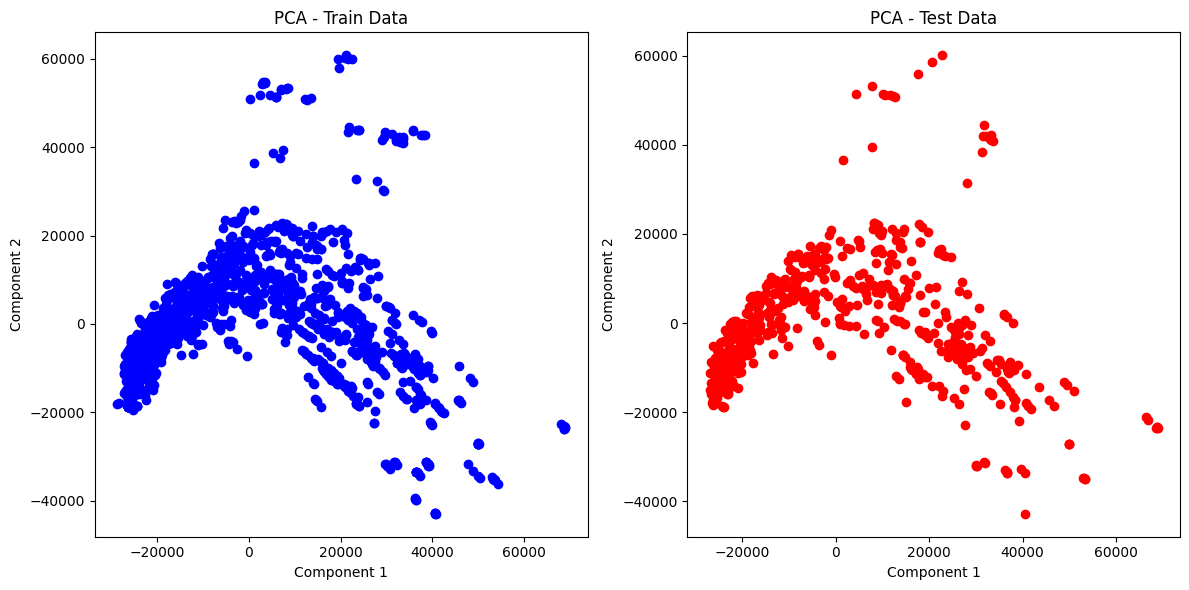}
\caption{PCA - Train \& Test for Bangla-BERT} \label{bangla_bert}

\end{figure}
\subsection{Null Value Handling}

A rigorous check for null values was conducted post-tokenization. In instances where null values were encountered, we judiciously replaced them by adopting a strategy based on the maximum allowable token length.

\subsection{Batching}
For efficient model training, two distinct batch sizes were utilized. The first configuration involved a train batch size of 16 and a test batch size of 8, while the second configuration featured a train batch size of 32 and a test batch size of 16. This bifurcation aimed to explore the impact of batch size on model performance.

\subsection{Model Training}
Three distinct models were chosen for the training phase: Bert-Base-Multilingual-Uncased\cite{bert_lr}, Bangla-Bert\cite{banglabert_lr}, and Roberta\cite{roberta_lr}. To attain optimal performance, these models underwent extensive training with varying epochs and learning rates. The long training process was used to guarantee an accurate evaluation of the models' capabilities and to deliver the highest level of accuracy possible.
In each iteration of the training loop, batches of tokenized question-context pairings were processed, producing projected answer spans. By comparing predicted answer locations with actual response placements, the loss was computed. Backpropagation gradients were used to adjust the model's parameters and reduce loss. At regular intervals, the system assessed its performance using validation data, and logging metrics like accuracy. Throughout training, we logged vital details including progress, losses, and metrics, offering insights into the learning process.

\section{Evaluation}
After fine-tuning, the system was evaluated using two key metrics: Exact Match (EM) and F1 score, which are commonly used in QA systems \cite{karra2024analysis}. The EM score measures the percentage of predicted answers that exactly match the ground truth answers. The F1 score balances precision and recall by considering partial overlaps between predicted and correct answers. These metrics provide insight into both the system’s accuracy and its ability to handle variations in the way questions are phrased.

\[
\text{EM} = \frac{\text{Number of Correct Predictions}}{\text{Total Number of Questions}} \times 100
\]
\\
\[
\text{F1} = \frac{2 \times \text{Precision} \times \text{Recall}}{\text{Precision} + \text{Recall}}
\]

\section{Experimental Analysis}
This section presents the experimental setup, results of the
QA system, analyses of the system’s performance.

\subsection{Experimental Setup}
The proposed framework was developed on a single 64-bit PC with an AMD Ryzen 7 5800X 8-Core CPU running at 3.80GHz and 32 GB of RAM. Additionally, Google Colab has been used to implement the coding portion with T4 GPU for faster processing.

\subsection{Result and Analysis}

In this comprehensive analysis of model performance, we conducted an in-depth evaluation of three distinct models: BERT Base, RoBERTa Base, and Bangla BERT, using critical performance metrics such as F1 scores and Exact Match (EM) scores. Our aim was to understand the impact of several key hyperparameters and model features on their effectiveness in a specific task.

\begin{table}[htbp]
\caption{Models with Parameters and Accuracy}
\label{table: table1}
\begin{adjustbox}{center}
\begin{tabular}{|l|l|l|l|l|l|l|}
\hline

Model   & Batch  & Stop & Learn & Epoch &  F1 & Exact \\

Name   & Size &  Word &  Rate & Size & Score & Match \\
\hline
   & 16/8       & Yes       & 2 e-4         & 10    & 0.28       & 0.01       \\
             &            & Yes       &  2 e-5    &    20    & \textbf{0.63}       & \textbf{0.40}       \\
            \cline{2-7}
            
             BERT Base &       & Yes       & 2 e-4         & 10    & 0.55       & 0.30       \\
             &      32/16      & Yes       & 2 e-5         & 20    & \textbf{0.61}       & \textbf{0.37}       \\
             &            & No        & 2 e-5         & 20    & 0.61       & 0.37       \\
\hline
 & 16/8       & Yes       & 2 e-4         & 10    & 0.24       & 0.35       \\
             &            & Yes       & 2 e-5         & 20    & \textbf{0.40}       & \textbf{0.36}       \\
            \cline{2-7}
            RoBERTa Base &      & Yes       & 2 e-4         & 10    & 0.19       & 0.27       \\
             &          32/16  & Yes       & 2 e-5         & 10    & 0.34       & 0.30       \\
             &             & Yes       & 5 e-5         & 20    & \textbf{0.38}       & \textbf{0.32}       \\
             &            & No        & 2 e-5         & 20    & 0.42       & 0.28       \\
\hline
 & 16/8       & Yes       & 2 e-4         & 10   & \textbf{0.75}       & \textbf{0.53}       \\
            \cline{2-7}
           Bangla BERT   &       & Yes       & 2 e-4        & 10    & \textbf{0.75}       & \textbf{0.53}       \\
             &      32/16      & No        & 2 e-4         & 10    & 0.67       & 0.43       \\
\hline

\end{tabular}
\end{adjustbox}
\end{table}

Here [~\ref{table: table1}], the standout model in terms of performance was Bangla BERT. This model consistently delivered the highest F1 score (0.75) and EM score (0.53) when configured with a batch size of 16/8, the inclusion of stop words, a learning rate of 2e-4, and training for 10 epochs. These results underline the robustness and adaptability of Bangla BERT in handling the given task effectively. Conversely, RoBERTa Base continuously provided the worst results. When configured with a batch size of 32/16, the inclusion of stop words, a learning rate of 2e-4, and trained for 10 epochs, consistently produced the poorest results with the lowest F1 score (0.19) and EM score (0.27). This finding suggests that this setup was not satisfactory to the RoBERTa Base model. BERT Base, a well-known and widely-used model, indicated performance outcomes that fell inside a particular range when compared to RoBERTa Base and Bangla BERT. Notably, certain hyperparameters and configuration choices have a significant influence on how well BERT Base performs. With a batch size of 16/8, stop words included, a learning rate of 2e-5, and 20 training epochs, the highest F1 score obtained during our testing of BERT Base was 0.63. For this arrangement, the corresponding EM score was 0.40.
Intriguing variations in model performance when stop words were included were also shown by our findings. Stop words had no impact on the BERT basis, as evidenced by the constant F1 and EM scores of 0.61 and 0.37. With stop words, however, RoBERTa base's performance declined, with F1 dropping from 0.42 to 0.38. On the other hand, stop words enhanced the Bangla BERT basis's performance, raising F1 from 0.67 to 0.75 and EM from 0.43 to 0.53 respectively. These results emphasize the model- and language-specific effects of stop words in NLP tasks.

\begin{figure}
\centering
\includegraphics[width=250px]{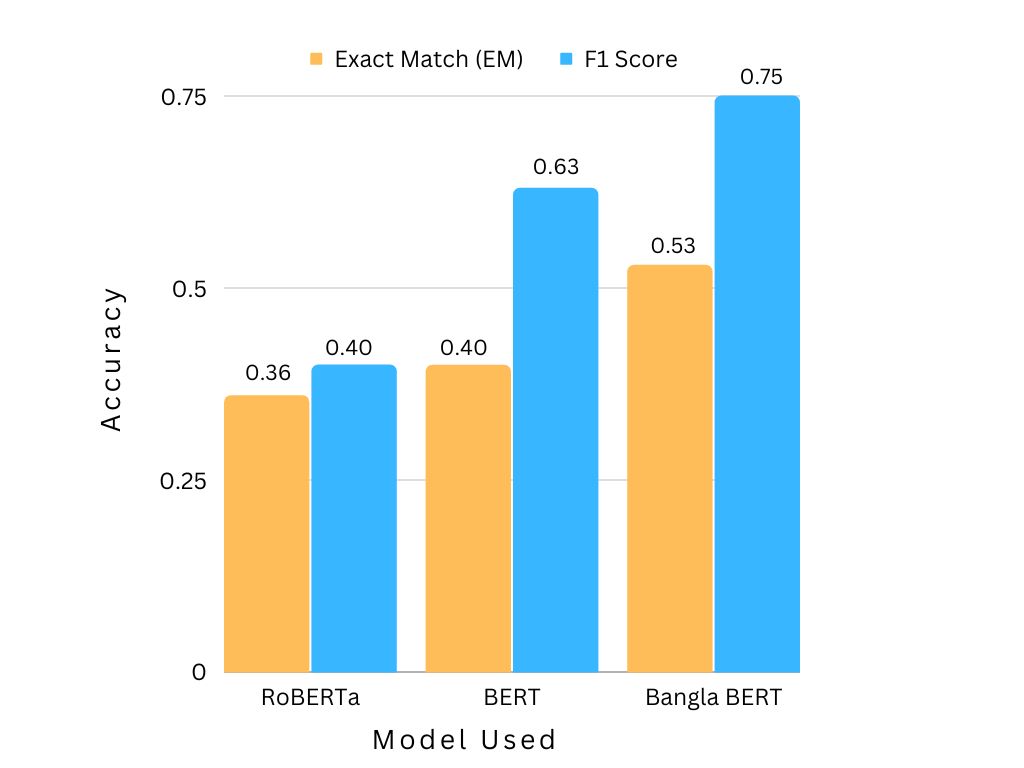}
\caption{Model's Accuracy Comparison} \label{EM}
\end{figure}

Additionally, the impact of batch size varied according to the model and additional hyperparameters. Bangla BERT benefited from a smaller batch size (16/8), but a larger one (32/16) was negative. This finding underscores the importance of fine-tuning batch size in conjunction with other hyperparameters to maximize model effectiveness. Furthermore, we observed that lower learning rates (2e-5) consistently outperformed higher rates.

However, the ideal learning rate may vary depending on the model architecture and other hyperparameters, emphasizing the need for careful tuning to achieve optimal performance. Also, our observations revealed that both the F1 and EM scores exhibited a consistent upward trend as the number of training epochs increased, up to a certain point.However, beyond this critical threshold, further increases in the number of epochs either led to a decline in scores or resulted in a plateau.

\section{Limitation}
In our study, it is essential to acknowledge several limitations that have influenced our research outcomes. Firstly, we worked with a relatively smaller dataset, which may have restricted the generalizability of our findings. Additionally, our dataset contained spelling mistakes originating from the source materials, primarily from NTCB books. While we attempted manual corrections, it was challenging to address all of them, potentially impacting the quality of the dataset. Another limitation pertains to the absence of proper answers in the passage. This issue arose because not all questions in our dataset were factoid-type questions, making it difficult to extract precise answers directly from the text. Furthermore, the repetition of passages within the dataset introduced a time-consuming aspect to our research, potentially affecting efficiency. We also encountered challenges in fitting the dataset seamlessly with different tokenizers like BERTTokenizer, RoBERTaTokenizer, and ElectraTokenizer, which could have introduced inconsistencies in our analysis. Lastly, computational limitations imposed constraints on the scale and complexity of our experiments. 

\section{Future Work}
Several promising avenues for future research emerge from our study's limitations. Firstly, there is an opportunity to develop specialized models for handling a broader range of question types, including Multiple Choice Questions (MCQs), Fill-in-the-blanks, and True-False questions. This diversification can lead to more versatile and robust natural language understanding systems.  Expanding our dataset is another critical direction, allowing for richer training and evaluation scenarios. To improve data quality, comprehensive efforts should focus on cleaning the dataset to eliminate spelling mistakes, ensuring that the model's performance is not hindered by inaccuracies originating from source materials. Additionally, exploring tokenization methods that better suit the dataset's characteristics can enhance the consistency and effectiveness of natural language processing tasks. Tailoring tokenizers to the specific nuances of the data is essential for optimizing model performance. Lastly, categorizing questions into different types can be a valuable future endeavour. By classifying questions based on their characteristics and information-seeking intent, we can tailor models and strategies to handle them more effectively. These future research directions have the potential to significantly advance the field of natural language understanding and improve the practical applications of language models.

\section{Conclusion}
In conclusion, our research endeavours have been dedicated to the meticulous development of a Bangla Question-Answering (QA) system, focusing on both dataset creation and model training. Our primary aim was to craft an effective and precise Bangla QA system capable of understanding the intricacies of the Bangla language. Central to our project is the construction of a customized dataset, comprising approximately 3,000 question-and-answer pairs. This dataset, thoughtfully curated with the assistance of human annotators who referenced NCTB textbooks, serves as the cornerstone of our research. It provides a rich contextual foundation for Bangla QA, with each passage offering substantial context for answering related questions. The careful collection of responses for diverse question types by human annotators ensures the dataset's reliability and relevance. In the realm of model training, we selected and rigorously trained three distinct models: Bert-Base-Multilingual-Uncased, Bangla-Bert, and Roberta. Through an iterative training process involving variations in epochs and learning rates, we sought to optimize model performance and gain insights into hyperparameter effects. Our performance analysis, using crucial metrics like F1 scores and Exact Match (EM) scores, identified Bangla BERT as the standout performer, consistently achieving the highest scores. Conversely, RoBERTa Base yielded suboptimal results under our specific configuration. These findings not only contribute to the development of Bangla QA systems but also pave the way for future research directions, including specialized models, dataset expansion, data quality enhancement, tailored tokenization, and question categorization, to further advance the field of natural language understanding in the Bangla language.

\section{Ethical Considerations}
We carefully checked and confirm that the work
conducted in this paper does not violate any ethical
considerations.

\bibliographystyle{IEEEtran}
\bibliography{refs.bib}

\begin{thebibliography}{10}
\providecommand{\url}[1]{#1}
\csname url@samestyle\endcsname
\providecommand{\newblock}{\relax}
\providecommand{\bibinfo}[2]{#2}
\providecommand{\BIBentrySTDinterwordspacing}{\spaceskip=0pt\relax}
\providecommand{\BIBentryALTinterwordstretchfactor}{4}
\providecommand{\BIBentryALTinterwordspacing}{\spaceskip=\fontdimen2\font plus
\BIBentryALTinterwordstretchfactor\fontdimen3\font minus
  \fontdimen4\font\relax}
\providecommand{\BIBforeignlanguage}[2]{{%
\expandafter\ifx\csname l@#1\endcsname\relax
\typeout{** WARNING: IEEEtran.bst: No hyphenation pattern has been}%
\typeout{** loaded for the language `#1'. Using the pattern for}%
\typeout{** the default language instead.}%
\else
\language=\csname l@#1\endcsname
\fi
#2}}
\providecommand{\BIBdecl}{\relax}
\BIBdecl

\bibitem{mervin2013overview}
R.~Mervin, ``An overview of question answering system,'' \emph{International
  Journal Of Research In Advance Technology In Engineering (IJRATE)}, vol.~1,
  pp. 11--14, 2013.

\bibitem{das2022banglaser}
R.~K. Das, N.~Islam, M.~R. Ahmed, S.~Islam, S.~Shatabda, and A.~M. Islam,
  ``Banglaser: A speech emotion recognition dataset for the bangla language,''
  \emph{Data in Brief}, vol.~42, p. 108091, 2022.

\bibitem{bert_lr}
J.~Devlin, M.-W. Chang, K.~Lee, and K.~Toutanova, ``Bert: Pre-training of deep
  bidirectional transformers for language understanding.'' in \emph{Proceedings
  of NAACL-HLT 2019}, vol.~1, no.~1, Jun. 2019, pp. 4171--4186.

\bibitem{banglabert_lr}
\BIBentryALTinterwordspacing
A.~Bhattacharjee, T.~Hasan, W.~Ahmad, K.~S. Mubasshir, M.~S. Islam, A.~Iqbal,
  M.~S. Rahman, and R.~Shahriyar, ``{B}angla{BERT}: Language model pretraining
  and benchmarks for low-resource language understanding evaluation in
  {B}angla,'' in \emph{Findings of the Association for Computational
  Linguistics: NAACL 2022}.\hskip 1em plus 0.5em minus 0.4em\relax Seattle,
  United States: Association for Computational Linguistics, Jul. 2022, pp.
  1318--1327. [Online]. Available:
  \url{https://aclanthology.org/2022.findings-naacl.98}
\BIBentrySTDinterwordspacing

\bibitem{roberta_lr}
Y.~Liu, M.~Ott, N.~Goyal, J.~Du, M.~Joshi, D.~Chen, O.~Levy, M.~Lewis,
  L.~Zettlemoyer, and V.~Stoyanov, ``Roberta: A robustly optimized bert
  pretraining approach.''\hskip 1em plus 0.5em minus 0.4em\relax arXiv preprint
  arXiv:1907.11692.

\bibitem{parsa_lr_1}
S.~Banerjee, S.~Kumar~Naskar, and S.~Bandyopadhyay, ``Reading comprehension
  based question answering system in bangla language with transformer-based
  learning,'' \emph{International Conference on Text, Speech and Dialogue},
  2014.

\bibitem{parsa_lr_2}
M.~Zhou, M.~Huang, and X.~Zhu, ``Robust reading comprehension with linguistic
  constraints via posterior regularization,'' \emph{IEEE/ACM Transactions on
  Audio, Speech, and Language Processing}, vol.~28, pp. 2500--2510, 2020.

\bibitem{tashik_lr_1}
F.~Bechet, E.~Antoine, J.~Auguste, and G.~Damnati, ``Question generation and
  answering for exploring digital humanities collections,'' in
  \emph{Proceedings of the Thirteenth Language Resources and Evaluation
  Conference}.\hskip 1em plus 0.5em minus 0.4em\relax Marseille, France:
  European Language Resources Association, Jun. 2022, pp. 4561--4568.

\bibitem{tashik_lr_2}
S.~M.~S. Ekram, A.~A. Rahman, M.~S. Altaf, M.~S. Islam, M.~M. Rahman, M.~M.
  Rahman, M.~A. Hossain, and A.~R.~M. Kamal, ``{B}angla{RQA}: A benchmark
  dataset for under-resourced {B}angla language reading comprehension-based
  question answering with diverse question-answer types,'' in \emph{Findings of
  the Association for Computational Linguistics: EMNLP 2022}.\hskip 1em plus
  0.5em minus 0.4em\relax Abu Dhabi, United Arab Emirates: Association for
  Computational Linguistics, Dec. 2022, pp. 2518--2532.

\bibitem{nihal_lr_1}
G.~Berger, T.~Rischewski, L.~Chiruzzo, and A.~Rosá, ``Generation of english
  question answer exercises from texts using transformers based models.''\hskip
  1em plus 0.5em minus 0.4em\relax IEEE, 2022.

\bibitem{nihal_lr_2}
T.~T. Mayeesha, A.~M. Sarwar, and R.~M. Rahman, ``Deep learning based question
  answering system in bengali,'' \emph{Journal of Information and
  Telecommunication}, vol.~5, no.~2, pp. 145--178, 2021.

\bibitem{abdullah_lr_1}
M.~Rathod, T.~Tu, and K.~Stasaski, ``Educational multi-question generation for
  reading comprehension,'' in \emph{Proceedings of the 17th Workshop on
  Innovative Use of NLP for Building Educational Applications (BEA
  2022)}.\hskip 1em plus 0.5em minus 0.4em\relax Seattle, Washington:
  Association for Computational Linguistics, Jul. 2022, pp. 216--223.

\bibitem{dataset_lr}
A.~Khondoker, E.~Ahmed, M.~I.~I. Tashik, S.~M.~I. Mahmud, and A.~F. Parsa,
  ``Textbook dataset from nctb”, mendeley data,'' vol.~V1.\hskip 1em plus
  0.5em minus 0.4em\relax Mendeley Data, 2023.

\bibitem{stopwords_bn_lr}
\BIBentryALTinterwordspacing
Genediazjr, ``Stopwords bengali (bn).'' [Online]. Available:
  \url{https://github.com/stopwords-iso/stopwords-bn}
\BIBentrySTDinterwordspacing

\bibitem{karra2024analysis}
R.~Karra and A.~Lasfar, ``Analysis of qa system behavior against context and
  question changes.'' \emph{Int. Arab J. Inf. Technol.}, vol.~21, no.~2, pp.
  191--200, 2024.

\end{thebibliography}

\end{document}